\documentclass[10pt,twocolumn,letterpaper]{article}
\pdfoutput=1

\usepackage{iccv}
\usepackage{times}
\usepackage{epsfig}
\usepackage{graphicx}
\usepackage{amsmath}
\usepackage{amssymb}
\usepackage{booktabs}
\usepackage{threeparttablex}
\usepackage{xspace}
\usepackage{multirow}
\usepackage{bbding}
\usepackage{appendix}
\graphicspath{{figures/}}
\usepackage{color}
\usepackage{float}
\usepackage[caption=false]{subfig}
\usepackage{listings}
\usepackage{colortbl}
\usepackage{enumitem}
\usepackage{booktabs}

\usepackage[breaklinks=true,bookmarks=false,colorlinks]{hyperref}

\usepackage{marvosym}

\newcommand\blfootnote[1]{%
  \begingroup
  \renewcommand\thefootnote{}\footnote{#1}%
  \addtocounter{footnote}{-1}%
  \endgroup
}

\newcolumntype{x}[1]{>{\centering\arraybackslash}p{#1pt}}

\newlength\savewidth
\newcommand{\tablestyle}[2]{\setlength{\tabcolsep}{#1}\renewcommand{\arraystretch}{#2}\centering\footnotesize}

\iccvfinalcopy
		
\def\iccvPaperID{9731} 

\begin{document}

\title{\ MGSampler: An Explainable Sampling Strategy for Video Action Recognition}

\author{
    Yuan Zhi \quad Zhan Tong \quad Limin Wang\textsuperscript{\Letter} \quad Gangshan Wu \\
State Key Laboratory for Novel Software Technology, Nanjing University, China\\
{\tt\small \{yuanzhi,tongzhan\}@smail.nju.edu.cn, \tt\small\{lmwang,gswu\}@nju.edu.cn}
}

\maketitle
\pagestyle{plain}


\begin{abstract}
Frame sampling is a fundamental problem in video action recognition due to the essential redundancy in time and limited computation resources. The existing sampling strategy often employs a fixed frame selection and lacks the flexibility to deal with complex variations in videos. In this paper, we present a simple, sparse, and explainable frame sampler, termed as {\em Motion-Guided Sampler} (MGSampler). Our basic motivation is that motion is an important and universal signal that can drive us to adaptively select frames from videos. Accordingly, we propose two important properties in our MGSampler design: motion sensitive and motion uniform. First, we present two different motion representations to enable us to efficiently distinguish the motion-salient frames from the background. Then, we devise a motion-uniform sampling strategy based on the cumulative motion distribution to ensure the sampled frames evenly cover all the important segments with high motion salience. Our MGSampler yields a new principled and holistic sampling scheme, that could be incorporated into any existing video architecture. Experiments on five benchmarks demonstrate the effectiveness of our MGSampler over the previous fixed sampling strategies, and its generalization power across different backbones, video models, and datasets. The code is available at \url{https://github.com/MCG-NJU/MGSampler}.
\end{abstract}
\blfootnote{ \Letter: Corresponding author.}

\section{Introduction}

Video understanding is becoming more and more important in computer vision research as huge numbers of videos are captured and uploaded online. Human action recognition~\cite{simonyan2014two,tran2018closer,wang2016temporal,lin_ICCV2019_TSM} has witnessed great progress in the past few years by designing various deep convolutional networks in videos. The core effort has been devoted to obtaining compact yet effective video representations for efficient and robust recognition. Compared with static images, the extra time dimension requires us to devise a sophisticated temporal module equipped with high capacity and figure out an efficient inference strategy for fast processing. However, in addition to these modeling and computational issues, a more fundamental problem in video understanding is {\em sampling}. Due to the essential redundancy in time and as well limited computational budget in practice, it is unnecessary and also infeasible to feed the whole video for subsequent processing. How to sample a small subset of frames is very important for developing a practical video recognition system, but it still remains to be an unsolved problem.

\begin{figure*}[htbp]
    \begin{center}
    \includegraphics[width=1\textwidth]{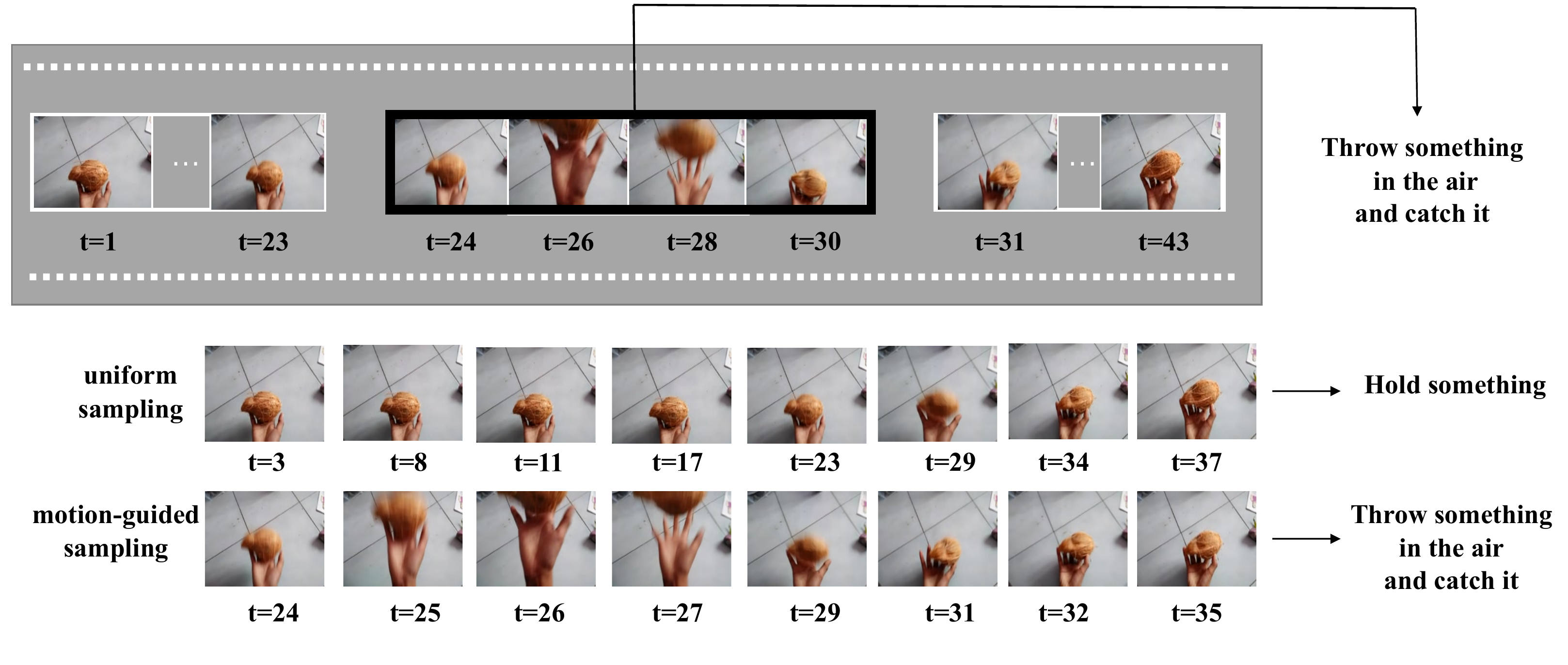}
    \end{center}
    \vspace{-8mm}
    \caption{Sample eight frames from a video of throwing something in the air and catching it. Due to the quick moment in action, uniform sampling may miss the key information while our sampling strategy can identify and select frames with large motion magnitude.}
    \label{pic1}
    \vspace{-5mm}
\end{figure*}

Currently, deep convolutional networks (CNNs) typically employ a fixed hand-crafted sampling strategy for training and testing in videos~\cite{simonyan2014two,tran2018closer,wang2016temporal}. In the training phase, CNN is trained on frames/clips which are randomly sampled either evenly or successively with a fixed stride from the original video. In the test phase, in order to cover the full temporal duration of video, clips are densely sampled from video and the final result is averaged from these dense prediction scores. There are multiple problems with these fixed sampling strategies. First, the action instance varies with different videos and sampling should not be fixed across videos. Second, not all the frames are of equal importance for classification and sampling should pay more attention to discriminative frames rather than irrelevant background frames. 

Recently, some works~\cite{wu2019adaframe,wu2019multi,fan2018watching} focus on frame selection in untrimmed videos, and try to improve the inference efficiency with an adaptive sampling module. These methods typically employ a learnable module to automatically select more discriminative frames for subsequent processing. However, these methods heavily rely on the training data with complicated learning strategies, and can not easily transfer to unseen action classes in practice. In addition, they typically deal with untrimmed video recognition by selecting foreground frames and removing background information. But it is unclear how to adapt them to trimmed video sampling due to the inherent difference between trimmed and untrimmed videos.

Based on the analysis above, how to devise a principled and adaptive sampling strategy for trimmed videos still needs further consideration in research. In this paper, we aim to present a simple, sparse and explainable sampling strategy for trimmed video action recognition, which is independent of the training data for good generalization ability and also capable of dealing with various video content adaptively. Our basic observation is that motion is a universal and transferable signal that can guide us to sample discriminative frames, in the sense that action related frames should be of high motion salience to convey most information about human movement while background frames typically contain no or limited irrelevant motion information. According to this motion prior, we can roughly analyze the frame importance and group frames into several segments according to their temporal variations. Consequently, these temporal segments enable us to perform a holistic and adaptive sampling to capture most motion information while suppressing the irrelevant background distraction, yielding a general frame sampler (MGSampler).

Specifically, to implement our motion-guided sampling, two critical components are proposed to handle motion estimation and temporal sampling, respectively. For motion representation, we use temporal difference at different levels to approximate human movement information for efficiency. In practice, temporal difference is highly correlated with motion information, and the absolute value of the difference is able to reflect the motion magnitude to some extent. For temporal sampling, based on the motion distribution along time, we present a uniform grouping strategy, where each segment should convey the same amount of motion salience. Then, according to this uniform grouping, we can perform adaptive sampling over the entire video by randomly picking a representative frame from each segment.  Figure~\ref{pic1} exhibits a vivid example of sampling frames from a video of the class ``throwing something in the air and catching it''. The motion relevant content only contains a small portion of the whole video (e.g.,  from the 24th frame to the 30th frame). If we use traditional uniform sampling, the important information between the 24th frame and 30th frame will be missed and as a result, the video is classified into holding a ball by mistake. In contrast, our motion-guided sampler selects more frames between the 24th frame and the 30th frame and makes a correct prediction. 

We conduct extensive experiments on five different trimmed video datasets: Something-Something V1 \& V2~\cite{sthsth}, UCF101~\cite{ucf101}, Jester~\cite{jester}, HMDB51~\cite{hmdb51}, and Diving48~\cite{diving}. Significant improvement is obtained on these datasets by adopting our motion-guided sampling strategy. It is worth noting that using the motion-guided sampling strategy will not increase the burden of computation and running time greatly. In addition, the method is agnostic to the network architecture, and can be used in both training and test phases, demonstrating its strong applicability.
\section{Related Work}
\begin{figure*}[ht]
    \begin{center}
    \includegraphics[width=1\textwidth]{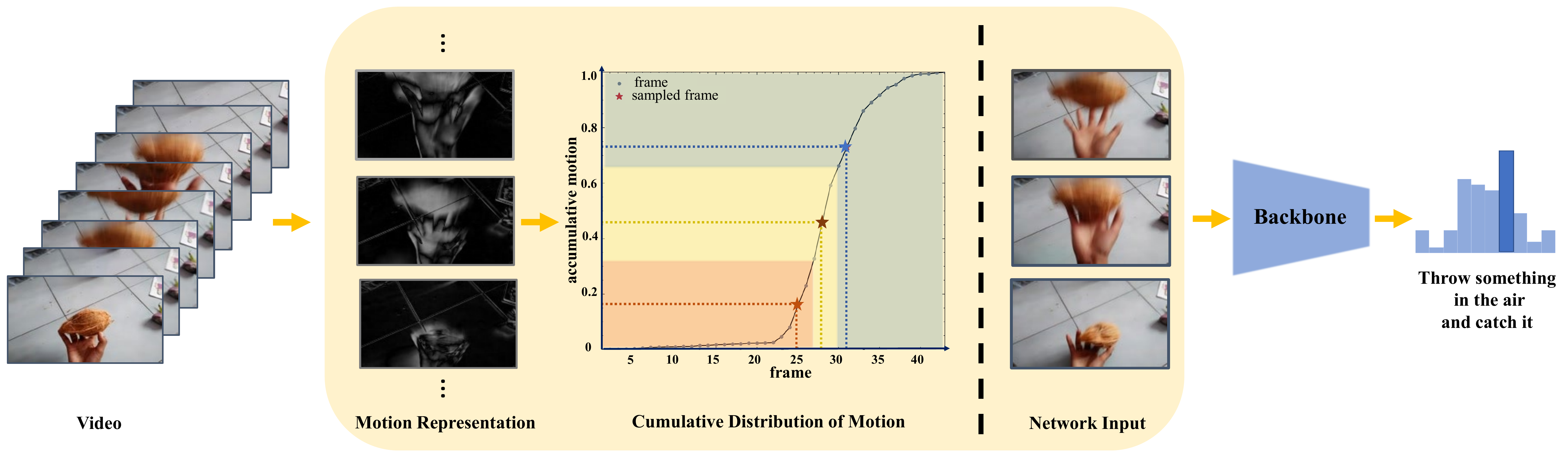}
    \end{center}
    \vspace{-3mm}
    \caption{\small {\bf Motion-guided Sampler (MGSampler).} Our MGSampler aims to on-the-fly select frames containing rich motion information to help the classifier see the whole process of action. Our proposed MGSampler is a general and flexible sampling scheme, that could be easily deployed for any existing video models for action recognition. }
    \label{fig:framework}
    \vspace{-5mm}
\end{figure*}

\paragraph{\bf Action Recognition.} Action recognition is a task to identify various human actions in a video. The last decade has witnessed a growing research interest in video action recognition with the availability of large-scale datasets and the rapid progress in deep learning. Methods can be generally categorized into four types: (1) Two-Stream Networks or variants: One stream takes RGB images as input to model appearance and another stream takes optical flow as input to model motion information. In the prediction stage, scores from two streams were averaged in a late fusion way~\cite{simonyan2014two}. Based on this architecture, several works were proposed for a better fusion of two streams~\cite{feichtenhofer2016convolutional,wang2016two}. (2) 3D CNNs: 3D CNN for action recognition aims to learn features along both spatial and temporal dimensions~\cite{ji20123d,tran2015learning,feichtenhofer_CVPR2020_X3D}. However, 3D CNN suffers from more computational cost than their 2D competitors due to the temporal dimension. In order to reduce computational costs, these works~\cite{tran2018closer,qiu2017learning} decomposed 3D convolution into a 2D convolution and a 1D temporal convolution or integrated 2D CNN into 3D CNN~\cite{zhou2018mict}. (3) Mixed spatiotemporal network models: ECO~\cite{zolfaghari_ECCV2018_ECO} and TSM~\cite{lin_ICCV2019_TSM} designed the lightweight models to fuse spatiotemporal features. MFNet~\cite{lee_ECCV2018_motionFilter}, TEINet~\cite{liu_AAAI2020_TEINet}, TEA~\cite{li2020tea}, MSNet~\cite{kwon2020motionsqueeze} and others~\cite{jiang_ICCV2019_STM,sudhakaran2020gate,weng2020temporal,wu2020MVFNet,wang2021tdn} explored better temporal modeling architecture for motion representation. (4) Long-term network models: short-term clip-based networks are unable to capture long-range temporal information. Several methods were proposed to overcome this limitation by stacking more frames with RNN~\cite{ng_CVPR2015_beyondLSTM} or long temporal convolution~\cite{varol_PAMI18_ltc}, or using a sparse sampling and aggregation strategy~\cite{wang2016temporal,zhou_ECCV2018_TRN,zhang_ICLR2020_V4D}. Unlike them, our goal is not designing a better model but devising effective frame sampling for a more fundamental issue in video analysis.

\vspace{-5mm}
\paragraph{\bf Frame Sampling.} For some 3D CNN based methods~\cite{tran2015learning,carreira2017quo, feichtenhofer_CVPR2020_X3D}, the video clip is obtained by choosing a random frame as the starting point. Then the next 64 consecutive frames in the video are subsampled uniformly to a fixed number of frames. TSN~\cite{wang2016temporal} performed a simple and effective sampling strategy where frames are uniformly sampled along the whole temporal dimension. The above two sampling strategies are commonly used by different models. However, they treated every frame equally and ignored the redundancy between frames, so selecting salient frames or clips conditioned on inputs is a key issue for efficient action recognition. Recently, several works proposed reinforcement learning (RL) to train agents with policy gradient methods to choose frames. FastForward~\cite{fan2018watching} utilized RL for both frame skipping planning and early stop decision making to reduce the computation burden for untrimmed video action recognition. Adaframe~\cite{wu2019adaframe} proposed a LSTM augmented with a global memory to search which frames to use over time, which was trained by policy gradient methods. Multi-agent~\cite{wu2019multi} uses N agents in the framework and each agent was responsible for selecting one informative frame/clip from an untrimmed video.  DSN~\cite{zheng2020dynamic} presented a dynamic version of TSN with RL-based sampling.
In order to avoid complex RL policy gradients, LiteEval~\cite{wu2019liteeval} proposed a coarse-to-fine and differentiable framework that contains a coarse LSTM and a fine LSTM organized hierarchically, as well as a gating module for selecting either coarse or fine features. AR-Net~\cite{meng2020ar} addressed both the selection of optimal frame resolutions and skipping in a unified framework and learned the whole framework in a fully differentiable manner. 
Audio has also been used as an efficient way to select salient frames for action recognition. SCSampler~\cite{korbar2019scsampler} used a lightweight CNN as the selector to sample clips at test time using salience scores. In order to train the selector effectively, they leveraged audio as an extra modality. Listen to Look~\cite{gao2020listen} used audio as a preview mechanism to eliminate both short-term and long-term visual redundancies for fast video-level recognition.
Though these approaches bring improvement in action recognition, their target is long and untrimmed videos rather than short and trimmed videos. What is more, the design of sampling module is usually complex and the training process requires large number of training samples with long training time. Instead, our goal is to present a simple, general, explainable frame sampling module without any learning strategy.

\section{Method}

In this section, we give detailed descriptions of our motion-guided sampling strategy. First, we give an overview of our motion-guided sampling. Then, we introduce the details of representing motion information of each frame. Finally, we elaborate on the concepts of using the cumulative distribution of motion magnitude to guide the sampling (MGSampler).
 
\subsection{Overview}
Videos are composed of a sequence of densely-captured frames. Due to temporal redundancy and limited computational resource, it is usual to sample a subset of frames to develop an efficient yet accurate action recognition method. Our proposed motion-guided sampling is a general and flexible module to compress the whole video into a fixed number of frames, which could be used for subsequent recognition with any kind of video recognition network (e.g., TSM, TEA, etc.). 

Motion prior is the core of our proposed sampling module and we assume this prior knowledge is general and transferable across videos and helpful to devise a universal sampler. Based on this assumption, we devise an adaptive sampling strategy with two important properties: {\em motion sensitive} and {\em motion uniform}. Concerning the requirement of motion sensitive, we hope our sampler is able to identify the motion salience along the temporal dimension and distinguish the action-relevant frames from the background. For the property of motion uniform, we expect our sampler can automatically select frames evenly according to the motion information distribution. In this sense, our sampled frames need to distribute uniformly over all of the temporal motion segments to cover the important details of action instances.  To accomplish the above requirements of motion-guided sampling, we design two critical components: {\em motion representation} and {\em motion-guided sampling}. For motion representation, to balance between accuracy and efficiency, we use the temporal difference to approximately capture the human movement. For motion-guided sampling, we devise a uniform sampling strategy based on the cumulative motion distribution to ensure cover all the important motion segments in the entire videos. Next, we will give detailed descriptions of these two components. 

\subsection{Motion Representation}
As RGB images usually represent static appearance at a specific time point, we need to consider the temporal variations of adjacent frames to leverage temporal context for motion estimation.

Optical flow~\cite{horn1981determining} is a common choice for motion representation, but the high computational cost makes it infeasible for efficient video recognition. Many works have been proposed to estimate optical flow with CNN ~\cite{dosovitskiy2015flownet,ilg2017flownet,fan2018end,ranjan2017optical} or explore alternatives of optical flow such as RGBdiff~\cite{wang2016temporal}, optical guided feature~\cite{sun2018optical}, dynamic image~\cite{bilen2017action} and fixed motion filter~\cite{lee2018motion}. Our target is to obtain an efficient yet relatively accurate motion representation to guide the subsequent sampling. We propose two motion representations based on different levels at little computation cost for selecting frames.

\begin{figure*}[ht]
    \begin{center}
    \includegraphics[width=1\textwidth]{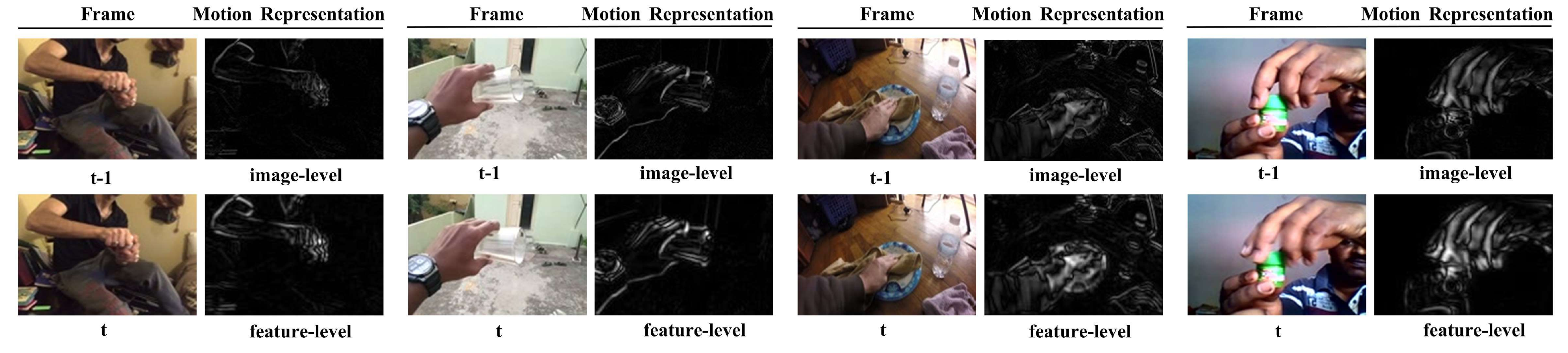}
    \end{center}
    \vspace{-3mm}
    \caption{Examples of original frames and its corresponding motion representation.The RGB frame contains rich appearance information and motion representation retains salient motion cues. Compared with image-level difference, feature-level difference captures more detailed and core motion displacement. }
    \label{2}
    \vspace{-2mm}
\end{figure*}

\paragraph{\bf Image-level Difference.} RGB difference between two consecutive frames describes the appearance change and has the correlation with the estimation of optical flow. Therefore, we adopt image-level difference between adjacent RGB frames as an alternative lightweight motion representation for the proposed sampling strategy. As shown in Figure~\ref{2}, the image-level difference between frames usually reserves only motion-specific features and suppresses the static background.

Formally, given a frame $I(x, y, t)$ from the video $\mathbf{V} \in \mathbb{R}^{T \times H \times W}$ where $T, H, W$ is the length, height, and width of the video, to formulate its motion magnitude, we first subtract each pixel value of the previous frame $I(x,y,t-1)$ from the current frame $I(x,y,t)$, then accumulate the absolute value of difference values over the spatial domain for each frame: 
\begin{equation}
\small
S_{t}=\sum_{y=1}^{H} \sum_{x=1}^{W}|I(x, y, t)- I(x, y, t-1)|, t \in\{2,3{},\ldots, T\}\label{eq1}
\end{equation}
where $S_{t}$ describes the motion signal of frame $I(x, y, t)$ and $S_{1}=0$. We further normalize $S_t$ with $\ell_1$-norm to obtain motion salience distribution $M_t$ (i.e., $\sum_t^T M_t=1$).
\vspace{-3mm}
\paragraph{\bf Feature-level Difference.} Although difference between original images can reflect motion information to some extent, more precise patterns such as motion boundaries and textures are hard to capture only by image-level difference. 

It is a consensus that convolution has the ability to extract feature and  filters in low convolutional layers usually describe boundaries and textures, whereas filters in high convolutional layers are more likely to represent abstract parts.  We reemphasize that the main idea in designing motion representation is to achieve a balance between computation and efficiency, so we perform shallow-layer convolution operation on original images. Then, in order to focus on small motion displacements and motion boundaries, we extend the subtraction operation to the feature space by replacing the original image $I(x, y, t)$ by its corresponding feature maps $F(x, y, t)$. The feature-level difference is defined as follows:
\begin{equation}
Diff_{i}(x, y, t)=F_{i}(x, y, t)-F_{i}(x, y, t-1)
\end{equation}
where the subscript $i \in\{1,2{},\ldots, C\}$ represents the $i$-th feature map of the original image and $C$ is the number of channel. In experiments, we use one convolutional layer which consists of eight 7$\times$7 convolutions with stride=1 and padding=3, following the design of PA module~\cite{zhang2019pan}. The padding operation avoids the reduction in spatial resolution.

Because $Diff(x, y, t)\in \mathbb{R}^{H \times W \times C}$ is three-dimensional, to formulate the feature-level difference, C channels are accumulated to 1 channel by the square sum operation, leading $Diff(x, y, t)\in \mathbb{R}^{H \times W}$. Then all pixel values are added into one value. The mapping $\mathbb{R}^{H \times W \times C} \rightarrow \mathbb{R}$ makes $Diff(x, y, t)$ represent the motion magnitude of each frame. 
\begin{equation}
\small
S_{t}=\sum_{y=1}^{H} \sum_{x=1}^{W}\sqrt{\sum_{i=1}^{C}\left(Diff_{i}(x, y, t)\right)^{2}}, t \in\{2,3{},\ldots, T\}
\end{equation}

Like image-level difference, we further normalize $S_t$ with $\ell_1$-norm to obtain motion salience distribution $M_t$, that is $\sum_t^T M_t=1$.

\subsection{Motion-Guided Sampling (MGSampler)}

\begin{figure}
   \centering
    \includegraphics[width=1.0\linewidth]{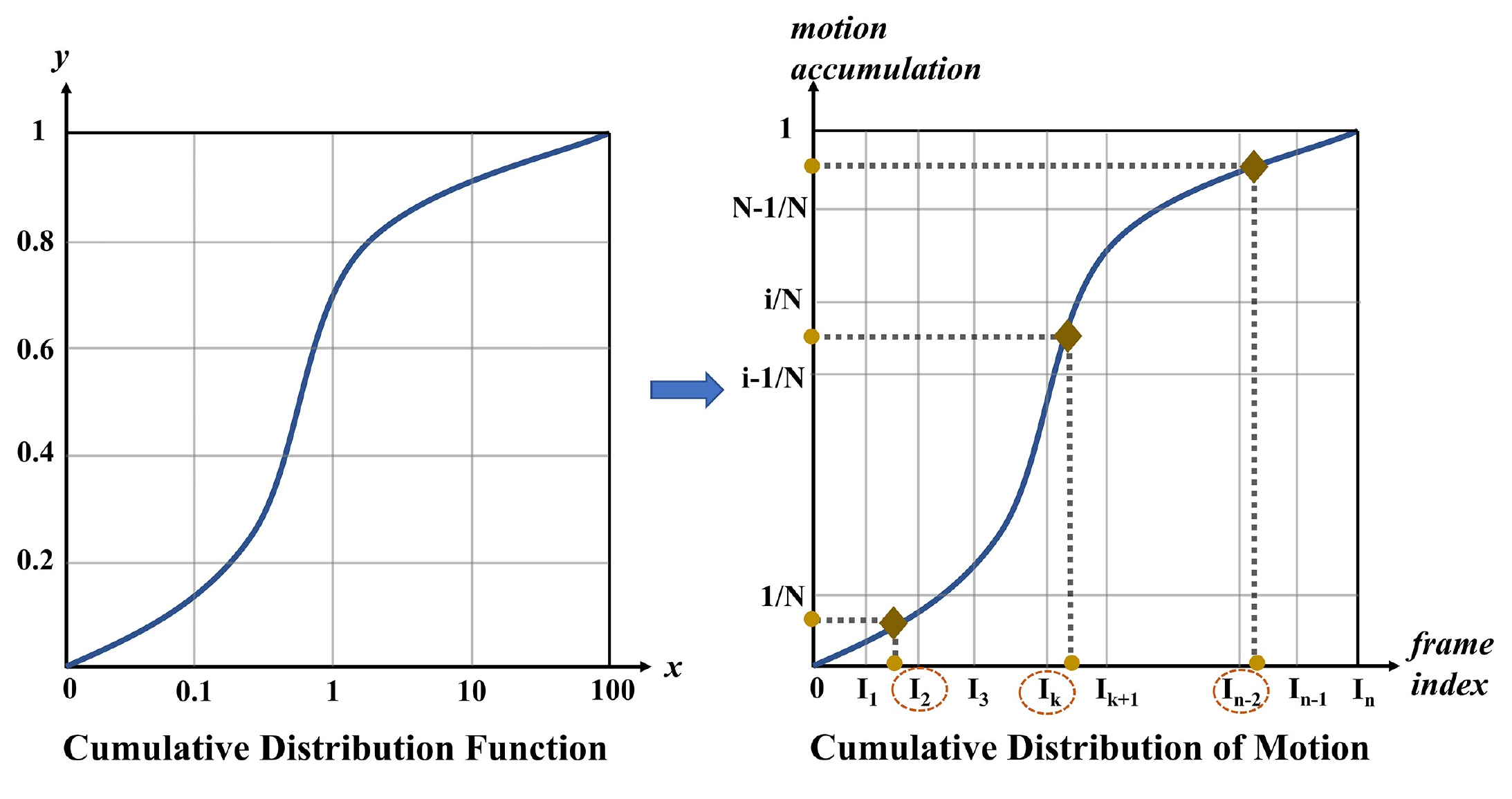}
    \vspace{-4mm}
    \caption{Inspired by the cumulative distribution function.}
    \label{fig:motivation}
    \vspace{-3mm}
\end{figure}

After obtaining the motion salience distribution along time $M_t$, we are ready to describe how to use it to perform motion-guided sampling. Similar to segment-based sampling in TSN~\cite{wang2016temporal}, our sampling is a holistic and duration-invariant strategy, in the sense that we sample over the entire video and compress the whole video into a subset of frames. In contrast to TSN which is a fixed sampling strategy, our motion-guided sampling adaptively selects frames according to motion uniform property and hopes the sampled frames could cover the important motion segments. In order to perform sampling adaptively according to motion distribution, we present a temporal segmentation scheme based on the cumulative motion distribution and then randomly sample a representative from each segment.

\begin{figure}
    \centering
    \includegraphics[width=0.6\linewidth]{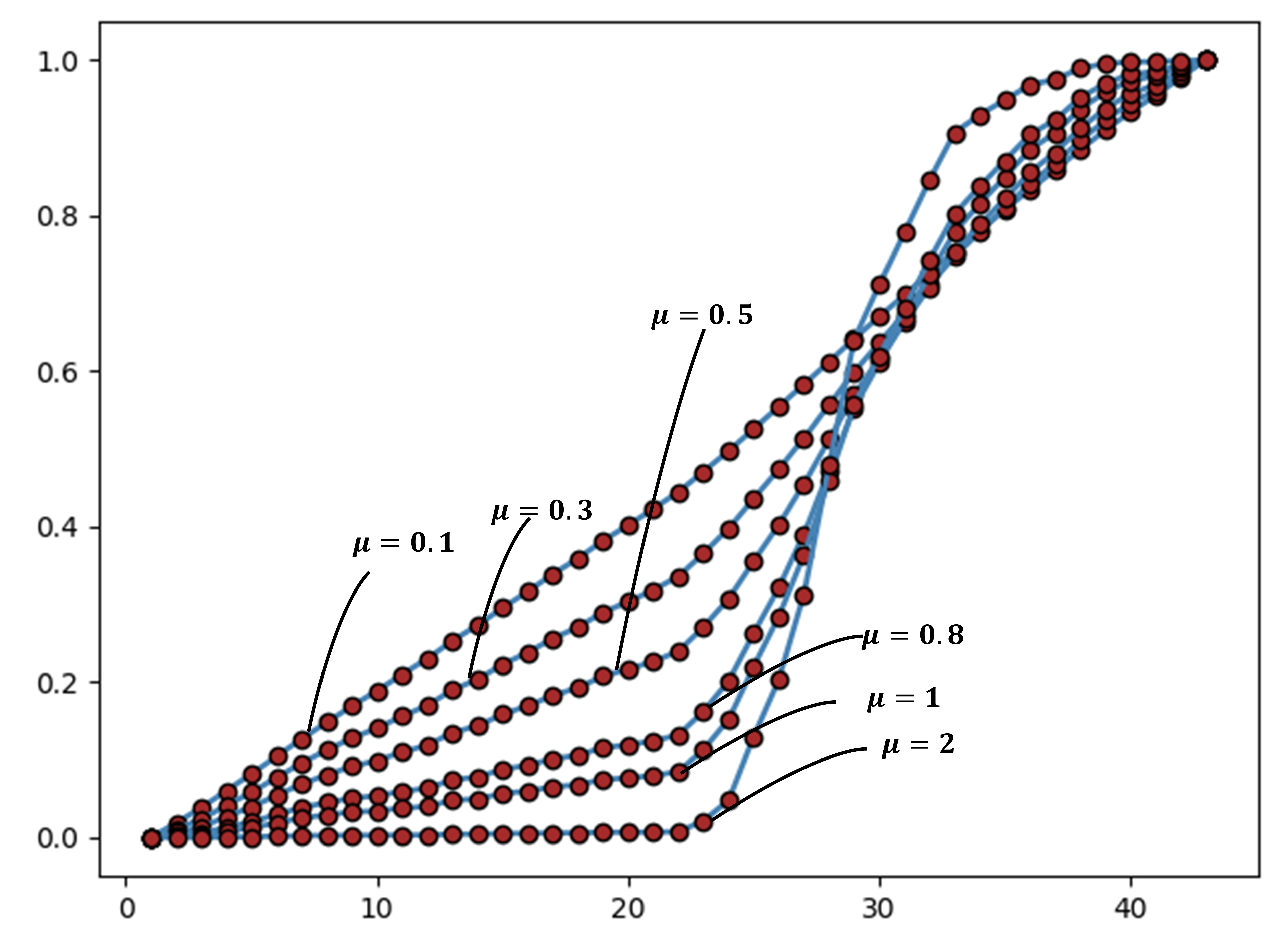}
    \vspace{-0.1mm}
    \caption{the cumulative motion distribution under different values of $\mu$.}
    \label{pic4}
    \vspace{-5mm}
\end{figure}

Specifically, the cumulative distribution function of a purely discrete variable $X$, having $n$ values $x_{1}, x_{2}, \ldots x_{n}$ with probability $p_{i}=p\left(x_{i}\right)$ is the defined by the following function: 
\begin{equation}
\small
F_{X}(x)=\mathrm{P}(X \leq x)=\sum_{x_{i} \leq x} \mathrm{P}\left(X=x_{i}\right)=\sum_{x_{i} \leq x} p\left(x_{i}\right),
\end{equation}
where $F_{X}$ is the accumulation of the probability from $x_{1}$ to $x_{n}$ and ranges from 0 to 1. Furthermore, the cumulative distribution function is non-decreasing and right-continuous.
\begin{equation}
F_{X}(x_{0})=0, \quad  F_{X}(x_{n})=1.
\end{equation}
Based on this definition of cumulative distribution function, we construct motion cumulative curve along the temporal dimension as shown in Figure~\ref{fig:motivation}, where x-axis represents the frame index and y-axis represents the motion information accumulation up to current frame. To further control the smoothness of motion-guided sampling, we introduce a hyper-parameter $\mu$ to adjust the original motion distribution $M_t$:
\begin{equation}
\hat{M}_{t}=\frac{\left(M_{t}\right)^{\mu}}{\sum_{t=1}^{T}\left(M_{t}\right)^{\mu}}.
\end{equation}
As shown in Figure~\ref{pic4}, a lower value for $\mu$ produces a more uniform probability distribution of motion magnitude. 

\begin{figure}[!t]
\centering
\subfloat[a typical distribution of motion magnitude of Sth-Sth V1.\label{tbl:ablation_diff}]{
\includegraphics[width=7cm]{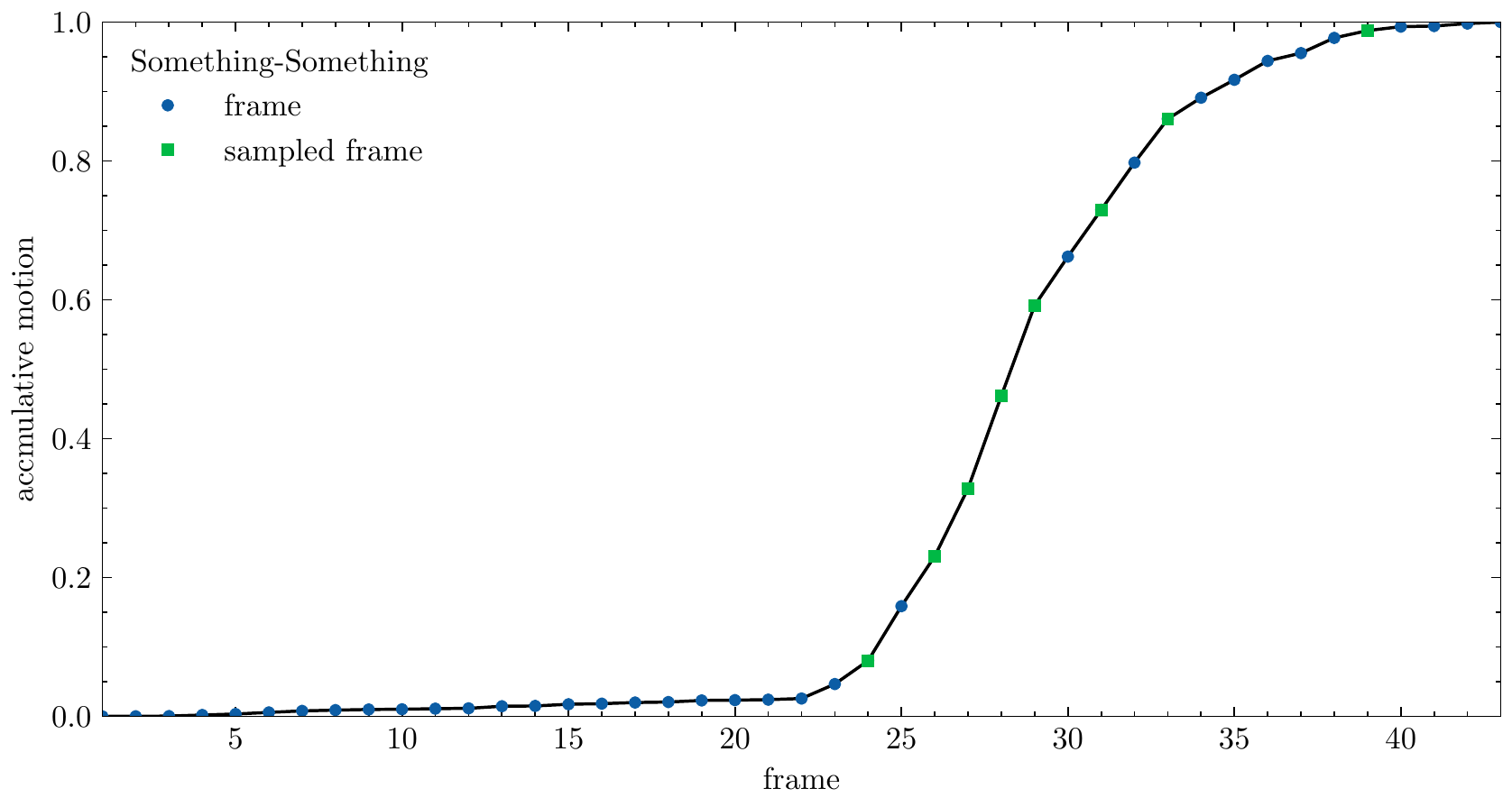}}
\vspace{-1mm}
\centering
\subfloat[a typical distribution of motion magnitude of Diving48.\label{tbl:adiff}]{
\includegraphics[width=7cm]{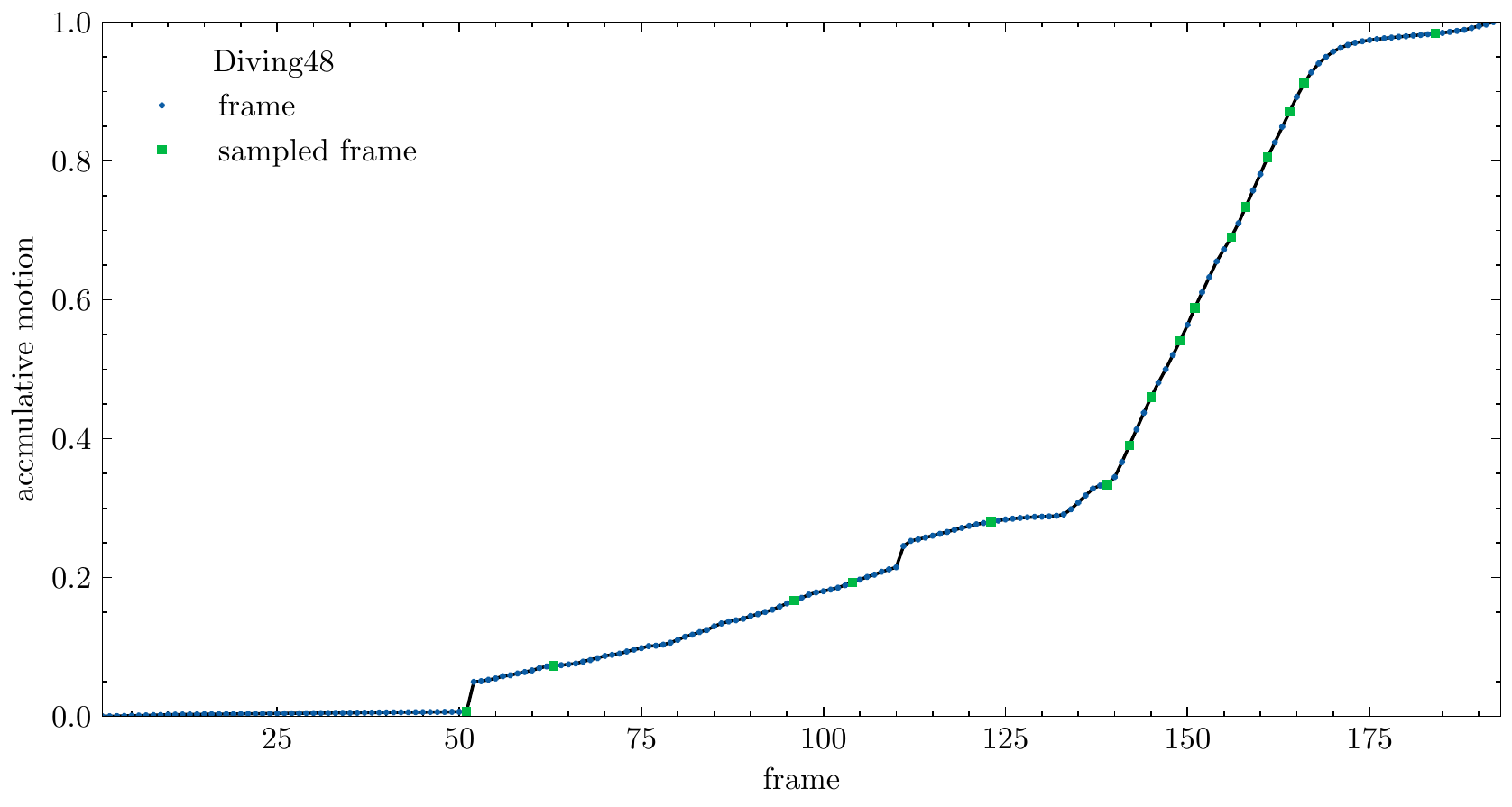}}
\vspace{1mm}
\caption{Different datasets has different video time and action categories, yet motion-guided sampling method can guide the sampling with the cumulative distribution of motion magnitude.}
\vspace{-6mm}
\label{fig:pic5}
\end{figure}

According to the obtained motion cumulative distribution curve, we now can perform our motion-guided sampling strategy. In order to sample N frames from the original video, the interval of y-axis is divided into $N$ parts evenly:$\left(\left[0, \frac{1}{N}\right],\left(\frac{1}{N}, \frac{2}{N}\right],\left(\frac{2}{N}, \frac{3}{N}\right], \ldots\left(\frac{N-1}{N}, \frac{N}{N}\right]\right)$. From each interval $\left(\frac{i-1}{N}, \frac{i}{N}\right]$, one value will be chosen randomly in its segment and its corresponding frame index on the x-axis will be picked out based on the curve. Considering that the x-axis value on the curve might not be an integer, we choose the integer closest to that value. Our sampling strategy is able to sample more frames during motion salient segments while sampling very small frames on static ones, thus allowing the subsequent video recognition models to focus on discriminative motion information learning. The sampled frames constitute a frame volume and will be fed into video CNNs to perform action recognition. In practice, we experiment with multiple network architectures and datasets Figure~\ref{fig:pic5} to verify the effectiveness of our motion-guided sampler.

\subsection{Discussion}
We have noticed that there are several sampling methods proposed recently, such as SCSampler~\cite{korbar2019scsampler}, DSN~\cite{zheng2020dynamic},  Adaframe~\cite{wu2019adaframe}, Listen to Look~\cite{gao2020listen} and AR-Net~\cite{meng2020ar}. However, their focus is completely different from ours. Firstly, they aim at sampling a reduced set of clips from namely long and frequently sparse videos with a typical length of a minute or more, while our target is to choose a more effective input with a fixed length in trimmed videos. Secondly, some of the methods need extra input to train the sampler. SCSampler~\cite{korbar2019scsampler} and Listen to Look~\cite{gao2020listen} use audio as an extra modality for exploiting the inherent semantic correlation between audio and the visual image. Thirdly, when training the sampler, Reinforcement Learning is commonly used where one agent or multiple agents are trained with policy gradient methods to select relevant video frames(Adaframe~\cite{wu2019adaframe}, DSN~\cite{zheng2020dynamic}). AR-net~\cite{meng2020ar} contains a policy network with
a lightweight feature extractor and an LSTM module. Both of the above training processes are complex and bring the network much extra computation.

In contrast to previous work, our proposed sampling strategy differs in three aspects. (1) The frame selection aims at selecting more effective frames with a fixed length in trimmed videos. (2) The sampling process doesn't need any extra input, making the input the same as the original. (3) MGSampler avoids complex training and is flexible enough to be inserted into other models.

\section{Experiments}
\subsection{Datasets and Implementation Details}
\paragraph{\bf Datasets.} We evaluate the motion-guided sampling strategy on five video datasets. These datasets can be grouped into two categories. (1) {\bf Motion-related datasets}: Something-Something V1\&V2~\cite{sthsth}, Diving 48~\cite{diving}, and Jester~\cite{jester}. For these datasets, motion information rather than static appearance is the key to action understanding. In Something-Something V1\&V2~\cite{sthsth}, the same action is performed with different objects (“something”) so that models are forced to understand the basic actions instead of recognizing the objects. It includes about 100K videos covering 174 classes. Jester~\cite{jester} is a collection of labeled video clips that show humans performing hand gestures in front of a laptop camera or webcam, containing 148k videos and 27 classes. Diving48~\cite{diving} is designed to reduce the bias of scene and object context in action recognition. It has a fine-grained taxonomy covering 48 different types of diving with 18K videos in total. (2) {\bf Scene-related datasets}: UCF101~\cite{ucf101} and HMDB51~\cite{hmdb51}. Action recognition in these datasets can be greatly influenced by the scene context. UCF101~\cite{ucf101} consists of 13,320 manually labeled videos from 101 action categories. HMDB51~\cite{hmdb51} is collected from various sources, e.g., web videos and movies, which proves to be realistic and challenging. It consists of 6,766 manually labeled clips from 51 categories. \vspace{-4mm}

\paragraph{\bf Implementation details.}
 In experiments, we use different models and backbones to verify the robustness of the  motion-guided sampling strategy. Experiments are conducted on MMAction2~\cite{contributors2020openmmlab}. For fair comparison, all the settings are kept the same during training and testing. Taking Sth-Sth V1 dataset and TSM model as an example, we utilize 2D ResNet pre-trained on ImageNet dataset as the backbone. During training, random scaling and corner cropping are utilized for data augmentation, and the cropped region is resized to $224\times224$ for each frame. The batch size, initial learning rate, weight decay, and dropout rate are set to 64, 0.01, 5e-4, and 0.5 respectively. The networks are trained for 50 epochs using stochastic gradient descent (SGD), and the learning rate is decreased by a factor of 10 at 20 and 40 epochs. During testing, 1 clip with T frames is sampled from the video. Each frame is resized to $256\times256$ , and a central region of size $224\times224$  is cropped for action prediction. The implementation on other backbones and datasets is similar to this setting.

\subsection{Ablation Studies}
\begin{table}[t]
\tablestyle{1.6pt}{1.2}
\setlength{\tabcolsep}{1mm}
\begin{center}
\begin{tabular}{ccccccccc}
\toprule
\textbf{}&{$\mu$} & \textbf{0}& \textbf{0.1} & \textbf{0.3} & \textbf{0.5} & \textbf{0.8} & \textbf{1} & \textbf{2}  \\ 
\toprule
\multirow{2}*{\textbf{Image Diff}} & Sth V1 & 45.6 & 46.2  & 46.6 & \bf{47.1}  & 46.5 &45.7 & 42.8 \\
\cline{2-9}
~ & Sth V2 & 57.9 & 58.2  & 59.7 & \bf{59.8} & 59.4 & 58.0 & 56.2 \\ \midrule
\multirow{2}*{\textbf{Feature Diff}} & Sth V1 & 46.0 & 46.5  & 46.8 & \bf{47.3}  & 46.7 &46.2 & 43.9 \\ \cline{2-9}
~ & Sth V2 & 58.2 & 58.5  & 60.0 & \bf{60.1} & 59.8 & 58.3 & 56.4 \\ 
\bottomrule
\end{tabular}
\end{center}
\vspace{-1mm}
\caption{The effect of different values of $\mu$ on the results of Something-Something V1\&V2. We use TSM model for the abalation study. Both train and test phase sample one clip of eight frames.}
\label{tab:smoothing}
\vspace{-2mm}
\end{table}

\begin{table}[t]
\tablestyle{1.6pt}{1.0}
\begin{center}
\begin{tabular}{x{40}x{30}x{55}x{55}}
\toprule
\textbf{Dataset} & \textbf{Original} & \textbf{Image Diff} & \textbf{Feature Diff}  \\
\toprule
\textbf{Sth-V1}  & 45.6 & 47.1(+1.5)  & 47.3(\textbf{+1.7})    \\ \midrule
\textbf{Sth-V2} & 57.9  & 59.8(+1.9) & 60.1(\textbf{+2.2})     \\ \midrule
\textbf{Diving-48}  & 35.2 & 36.9(+1.7) & 37.4(\textbf{+2.2})     \\ \midrule
\textbf{UCF-101}  & 94.5 & 94.9(+0.4)  & 95.2(\textbf{(+0.7})    \\ \midrule
\textbf{HMDB-51}  & 72.6 & 73.3(+0.7)  & 73.8(\textbf{+1.2})    \\ \midrule
 \textbf{Jester}  & 96.5 & 96.9(+0.4)  &  97.5(\textbf{+1.0})   \\ 
\bottomrule
\end{tabular}
\end{center}
\vspace{-1mm}
\caption{{\bf Performance of different motion representations.} The {\em original} means using TSN method to sample frames, which is the original sampling strategy in TSM. Noting that both UCF101 and HMDB51 have 3 splits, we report the average result on all splits.}
\label{tab:input_ablation}
\vspace{-5mm}
\end{table}

\paragraph{\bf Study on the smoothing hyperparameter.}
As shown in Figure~\ref{pic4}, the smoothing hyperparameter $\mu$ controls the smoothness degree in our motion-guided sampling. When $\mu$ equals 1, the motion magnitude maintains the original one. If $\mu$ is greater than 1, it will increase the difference of motion magnitude between frames. On the contrary, when $\mu$ is set to less than 1, the influence of motion is decreased and particularly if $\mu$ is 0, the sampling process is equivalent to TSN~\cite{wang2016temporal} method. 
We perform the ablation study on hyperparameter $\mu$ and the results are reported in Table~\ref{tab:smoothing}. We observe that $\mu = 0.5$ achieves the best result because it balances the relationship between the overall temporal structure and motion difference. We also observe that our motion-guided sampling is better than the baseline of TSN sampling (i.e., $\mu=0$) by around 1.5\% and 2\% on Sth V1 and V2.
\vspace{-4mm}
\paragraph{\bf Study on different motion representations.}
We design two motion representations based on different levels. The image-level difference is a quite convenient way to capture motion replacement, but it ignores some important features and motion boundaries. On the other hand, feature-level difference can represent more precise motion cue while it needs a little bit more computation. Considering that our goal is to find an efficient way to represent motion, we only add one shallow convolutional layer to the original input yet it brings significant improvement. PAN~\cite{zhang2019pan} indicates that when the convolutional layer goes deeper,  the performance based on feature-level difference degrades because high-level features have been highly abstracted and fail to reflect small motion replacement and boundaries.

To compare the performance based on the two motion representations, we conduct experiments on five different datasets, using TSM as the base model and inputting 8 frames. The result shows that regardless of the dataset, feature-level difference performs better than image-level difference mainly because differences in low-level features can capture small motion variations at boundaries.

\begin{table}[!t]
\centering
\tablestyle{1.6pt}{1.05}
  \footnotesize
  \begin{tabular}{lx{35}x{35}}  
    \toprule
    \textbf{Strategy}  & \textbf{Sth-V1} & \textbf{Sth-V2} \\
    \midrule
    Segment based sampling &  45.6 & 57.9\\
    Fixed stride sampling &  43.7  & 53.4\\
    Motion magnitude sampling     & 41.5  & 52.8\\
    \midrule
    \textbf{Motion-guided sampling(Ours)} & \bf{47.3} & \bf{60.1}\\
    \bottomrule
  \end{tabular}
  \vspace{2mm}
  \caption{Performance of different sampling strategies on the Something-Something V1 \& V2 dataset.}
  \label{tab:sampling}
  \vspace{-1mm}
\end{table}

\begin{table}[t]
\begin{center}
\footnotesize
\tablestyle{1.6pt}{1.05}
\begin{tabular}{x{42}ccx{32}x{45}}
\toprule
\textbf{Model} & \textbf{Backbone}& \textbf{Frames} & \textbf{TSN} & \textbf{MG Sampler}  \\ 
\toprule
TSM~\cite{lin_ICCV2019_TSM}  & ResNet50 & 8 & 45.6 & 47.1(\textbf{+1.5})  \\
TSM~\cite{lin_ICCV2019_TSM}  & ResNet50 & 16 & 47.2 & 48.6(\textbf{+1.4})  \\
TSM~\cite{lin_ICCV2019_TSM}  & ResNet101 & 8 & 46.9 & 47.8(\textbf{+0.9})  \\
TSM~\cite{lin_ICCV2019_TSM} & ResNet101 & 16 & 47.9 & 49.0(\textbf{+1.1})  \\
\midrule
TEA~\cite{li2020tea} & ResNet50 & 8 & 48.9 & 50.2(\textbf{+1.3})  \\
TEA~\cite{li2020tea} & ResNet50 & 16 & 51.9 & 52.9(\textbf{+1.0})  \\
TEA~\cite{li2020tea} & ResNet101 & 8 & 49.4 & 50.6(\textbf{+1.2})  \\
TEA~\cite{li2020tea} & ResNet101 & 16 & 52.0 & 53.2(\textbf{+1.2})  \\
\midrule
GSM~\cite{sudhakaran2020gate} & BNInception & 8 & 47.2 & 48.2(\textbf{+1.0})  \\
GSM~\cite{sudhakaran2020gate} & BNInception & 16 & 49.6 & 50.8(\textbf{+1.2})  \\
GSM~\cite{sudhakaran2020gate} & InceptionV3 & 8 & 49.0 & 50.1(\textbf{+1.1})  \\
GSM~\cite{sudhakaran2020gate} & InceptionV3 & 16 & 50.6 & 51.9(\textbf{+1.3})  \\
\bottomrule
\vspace{-2mm}
\end{tabular}
\caption{Motion-guided sampling improves the accuracy for all different backbones and models, proving to be quite robust. In this ablation experiment, we use image-level difference as motion representation.}
\vspace{-8mm}
\label{tab:ablation_backbone}
\end{center}
\end{table}

\begin{table}[!t]
\setlength{\tabcolsep}{1.8pt}
      \begin{center}
      \footnotesize
      \begin{tabular}{l c c c c  }
      \toprule
      \multicolumn{1}{c}{\bfseries Method} & \multicolumn{1}{c}{\bfseries Backbone} & \multicolumn{1}{c}{\bfseries Frames}   & \multicolumn{1}{c}{\begin{tabular}[c]{@{}c@{}}{\bfseries Sth-Sth V1} \\ {\bfseries Top-1 (\%)}\end{tabular}} & \multicolumn{1}{c}{\begin{tabular}[c]{@{}c@{}}{\bfseries Sth-Sth V2} \\ {\bfseries Top-1 (\%)}\end{tabular}}\\ 
      \midrule
      \midrule
      \multicolumn{1}{c}{I3D~\cite{carreira2017quo} } & \multicolumn{1}{c}{3D ResNet50} & \multirow{2}{*}{32$\times$3$\times$2} &  \multicolumn{1}{c}{41.6} & \multicolumn{1}{c}{-} \\
      \multicolumn{1}{c}{NL I3D~\cite{nonlocal}} & \multicolumn{1}{c}{3D ResNet50} &  & \multicolumn{1}{c}{44.4} & \multicolumn{1}{c}{-} \\
      \midrule
      \multicolumn{1}{c}{ECO~\cite{zolfaghari_ECCV2018_ECO}} & \multirow{2}{*}{BNIncep+R18} & \multicolumn{1}{c}{8$\times$1$\times$1}  & \multicolumn{1}{c}{39.6} & \multicolumn{1}{c}{-}  \\
      \multicolumn{1}{c}{ECO$_{En}$~\cite{zolfaghari_ECCV2018_ECO}} &   &  \multicolumn{1}{c}{92$\times$1$\times$1}  &   \multicolumn{1}{c}{46.4} & \multicolumn{1}{c}{-}\\
      \midrule
      \multicolumn{1}{c}{TSN~\cite{wang2016temporal} } & \multicolumn{1}{c}{BNInception} & \multirow{2}{*}{8$\times$1$\times$1} & \multicolumn{1}{c}{19.5} & \multicolumn{1}{c}{-} \\
      \multicolumn{1}{c}{TSN~\cite{wang2016temporal}  } & \multicolumn{1}{c}{ResNet50} &  &   \multicolumn{1}{c}{19.7} & \multicolumn{1}{c}{27.8} \\\midrule
      \multicolumn{1}{c}{TSM~\cite{lin_ICCV2019_TSM}   } & \multirow{2}{*}{ResNet50} & \multicolumn{1}{c}{8$\times$1$\times$1} &  \multicolumn{1}{c}{45.6} & \multicolumn{1}{c}{57.9} \\
      \multicolumn{1}{c}{TSM~\cite{lin_ICCV2019_TSM} } &  & \multicolumn{1}{c}{16$\times$1$\times$1} &  \multicolumn{1}{c}{47.2} & \multicolumn{1}{c}{59.9}\\
      \midrule
      \multicolumn{1}{c}{GST~\cite{luo2019grouped}   } & \multirow{2}{*}{ResNet50} & \multicolumn{1}{c}{8$\times$1$\times$1} &  \multicolumn{1}{c}{47.0} & \multicolumn{1}{c}{61.6} \\
      \multicolumn{1}{c}{GST~\cite{luo2019grouped} } &  & \multicolumn{1}{c}{16$\times$1$\times$1} &  \multicolumn{1}{c}{48.6} & \multicolumn{1}{c}{62.6}\\
      \midrule
      \multicolumn{1}{c}{TEINet~\cite{liu_AAAI2020_TEINet}} & \multirow{2}{*}{ResNet50} & \multicolumn{1}{c}{8$\times$1$\times$1} &  \multicolumn{1}{c}{47.4} & \multicolumn{1}{c}{61.3} \\
      \multicolumn{1}{c}{TEINet~\cite{liu_AAAI2020_TEINet}  } &  & \multicolumn{1}{c}{16$\times$1$\times$1}  & \multicolumn{1}{c}{49.9} & \multicolumn{1}{c}{62.1} \\
      \midrule
      \multicolumn{1}{c}{GSM~\cite{sudhakaran2020gate}} & \multirow{2}{*}{InceptionV3} & \multicolumn{1}{c}{8$\times$1$\times$1} &  \multicolumn{1}{c}{49.0} & \multicolumn{1}{c}{-} \\
      \multicolumn{1}{c}{GSM~\cite{sudhakaran2020gate}  } &  & \multicolumn{1}{c}{16$\times$1$\times$1}  & \multicolumn{1}{c}{50.6} & \multicolumn{1}{c}{-} \\
      \midrule
      \multicolumn{1}{c}{TDRL~\cite{weng2020temporal}} & \multirow{2}{*}{ResNet50} & \multicolumn{1}{c}{8$\times$1$\times$1} &  \multicolumn{1}{c}{49.8} & \multicolumn{1}{c}{62.6} \\
      \multicolumn{1}{c}{TDRL~\cite{weng2020temporal}  } &  & \multicolumn{1}{c}{16$\times$1$\times$1}  & \multicolumn{1}{c}{50.9} & \multicolumn{1}{c}{63.8} \\
      \midrule
      \multicolumn{1}{c}{MVFNet~\cite{wu2020MVFNet}} & \multirow{2}{*}{ResNet50} & \multicolumn{1}{c}{8$\times$1$\times$1} &  \multicolumn{1}{c}{48.8} & \multicolumn{1}{c}{60.8} \\
      \multicolumn{1}{c}{MVFNet~\cite{wu2020MVFNet}  } &  & \multicolumn{1}{c}{16$\times$1$\times$1}  & \multicolumn{1}{c}{51.0} & \multicolumn{1}{c}{62.9} \\
      \midrule

      \multicolumn{1}{c}{TEA~\cite{li2020tea} } & \multirow{2}{*}{ResNet50} & \multicolumn{1}{c}{8$\times$1$\times$1} &  \multicolumn{1}{c}{48.9} & \multicolumn{1}{c}{60.9}  \\
      \multicolumn{1}{c}{TEA~\cite{li2020tea}} &  & \multicolumn{1}{c}{16$\times$1$\times$1} &   \multicolumn{1}{c}{51.9} & \multicolumn{1}{c}{62.2} \\ 
      \midrule
      \midrule
      \multicolumn{1}{c}{MG-TEA(Ours)  } & \multirow{2}{*}{ResNet50} & \multicolumn{1}{c}{8$\times$1$\times$1} &  \multicolumn{1}{c}{50.4} & \multicolumn{1}{c}{62.5}  \\
      \multicolumn{1}{c}{MG-TEA(Ours) } &  & \multicolumn{1}{c}{16$\times$1$\times$1}  &   \multicolumn{1}{c}{53.2} & \multicolumn{1}{c}{63.8}  \\ 
      \multicolumn{1}{c}{MG-TEA(Ours)  } & \multirow{2}{*}{ResNet101} & \multicolumn{1}{c}{8$\times$1$\times$1} &  \multicolumn{1}{c}{50.8} & \multicolumn{1}{c}{63.7}  \\
      \multicolumn{1}{c}{MG-TEA(Ours) } &  & \multicolumn{1}{c}{16$\times$1$\times$1} &   \multicolumn{1}{c}{\bf{53.3}} & \multicolumn{1}{c}{\bf{64.8}}  \\

      \bottomrule
      \end{tabular}
      \footnotesize
     \vspace{-2mm}
      \end{center}
\caption{\label{tab:ss} \textbf{Comparison with other state-of-the-art methods on Something-Something V1\&V2.} We use TEA model with our motion-guided sampling strategy(MG-TEA) for the comparison. We mainly compare with other methods with similar backbones under the 1-clip and center crop setting. “-” indicates the numbers are not available for us. }
\vspace{-4mm}
\end{table}
\vspace{-4mm}
\paragraph{\bf Comparison with different sampling strategies.}
To better illustrate the effectiveness of our proposed motion-guided sampling, we compare it with three other sampling methods. First, we compare with two fixed sampling baselines: (1) segment based sampling~\cite{wang2016temporal} where 8 frames are sampled uniformly along the temporal dimension and (2) fixed stride sampling~\cite{carreira2017quo} where an 8-frame clip with a fixed stride (s=4) is randomly picked from the video. We see that our adaptive sampling module is better than those hand-crafted sampling schemes. Then, we compare with another adaptive sampling method based on motion magnitude (motion magnitude sampling), where 8 frames selected merely based on motion magnitude regardless of motion uniform assumption. We see that this alternative motion-guided sampling strategy yields much worse performance, which confirms the effectiveness of our strategy based cumulative motion distribution.
\vspace{-6mm}
\paragraph{\bf Varying backbones and models.}
We further demonstrate the robustness of our sampling strategy by varying the backbones and models. We choose ResNet 50~\cite{he_CVPR2016_resnet}, ResNet 101~\cite{he_CVPR2016_resnet}, BNInception~\cite{ioffe2015batch}, Inception V3~\cite{szegedy2016rethinking} for backbones and TSM~\cite{lin_ICCV2019_TSM}, TEA~\cite{li2020tea}, GSM~\cite{jiang_ICCV2019_STM} for models. Results on Table~\ref{tab:ablation_backbone} indicate that motion-guided sampling is able to bring consistent performance improvement across different methods.

\vspace{-5mm}
\paragraph{Efficiency and latency analysis.}
During training phase, we process the whole training set in advance by computing the difference. The 5$^{th}$ row of Table~\ref{tab:efficiency} reports the total computing time of processing training data. For testing, we first report the inference time of the standard sampling strategy(TSN) for each video in 6$^{th}$ row. Our MGSampler can slightly increase the inference time due to extra computation (7$^{th}$ row), which is acceptable. 
\subsection{Comparison with the state of the art}
We further report the performance of our motion-guided sampling on other datasets, including Diving48, UCF101, HMDB51, and Jester, and compare with the previous state-of-the-art methods. All the results are tested by one clip sampled from original video and reported in Table~\ref{tab:input_ablation}, Table~\ref{tab:ss} and Table~\ref{tab:diving48}. We see that our motion-guided sampling strategy is independent of the datasets and able to generalize well across datasets by bringing consistent performance improvement for different kinds of datasets with similar backbones under the single-clip and center-crop testing scheme.

\begin{table}[t]
\setlength{\tabcolsep}{1mm}
  \centering
  \tablestyle{1.6pt}{1.05}
  \footnotesize
  \begin{tabular}{lx{45}x{45}}  
    \toprule
    \textbf{Model}  & \textbf{Frames} & \textbf{Top-1} \\
    \midrule
    \midrule
    TSN \cite{wang2016temporal}     & 16    & 16.8\\
    C3D \cite{tran2015learning}     & 64  & 27.6\\
    R(2+1)D \cite{tran2018closer}     & 64  & 28.9\\
    \midrule
    P3D-ResNet50 \cite{qiu2017learning}  & 16 & 32.4 \\
    GST-ResNet50 \cite{luo2019grouped} & 16& 38.8 \\
    TEA-ResNet50 \cite{li2020tea} & 16& 36.0 \\
    GSM-InceptionV3\cite{sudhakaran2020gate} & 16 & 39.0\\
    \midrule
    \midrule
    MG-TEA-ResNet50(Ours)   & 16 & \textbf{39.5} \\
    \bottomrule
  \end{tabular}
  \vspace{2mm}
  \caption{Performance on the Diving-48 dataset compared with the state-of-the-art methods. For fair comparison, all the models are tested by one clip.}
  \label{tab:diving48}
  \vspace{-1mm}
\end{table}

\begin{table}[t]
\setlength{\tabcolsep}{3pt}

  \tablestyle{1.6pt}{1.05}
  \footnotesize
\resizebox{0.45\textwidth}{!}{
\begin{tabular}{cccccc}
\toprule
\textbf{} &\textbf{UCF101}& \textbf{HMBD51} & \textbf{Jester} & \textbf{Diving48}&\textbf{Sth-V2} \\
\toprule
\textbf{Training Set} & 9537 & 3750 & 118562 & 15943 & 168913 \\ 
\textbf{Testing Set}  & 3783 & 1530 & 14743 & 2096 & 24777 \\ 
\textbf{Average Frame}  & 187.3 & 96.6  & 36.0 & 159.6 & 45.8 \\ \midrule
\textbf{Training Time (all videos)}  & 72.4s & 23.2s  &  264.5s & 388.7s& 451.9s \\ \midrule
\textbf{TSN Sampling (each video)} & 6.5ms & 4.7ms &3.2ms & 6.8ms   & 4.4ms\\ \midrule
\textbf{MGSampler (each video)} & 6.9ms & 5.0ms & 3.5ms  & 7.4ms  & 5.0ms  \\
\bottomrule
\end{tabular}
}
\vspace{2mm}
\caption{Running time and latency of MGSampler.}
\label{tab:efficiency}
\vspace{-6mm}
\end{table}

\vspace{-1mm}
\section{Conclusion}
In this paper, we have presented a sparse, explainable, and adaptive sampling module for video action recognition, termed as MGSampler. Our new sampling module generally follows the assumption that motion is a universal and transferable prior information that enables us to design an effective frame selection scheme. Our motion-guided sampling shares two important ingredients: motion sensitive and motion uniform, where the former can help us identify the most salient segments against the background frames, and the latter enables our sampling to cover all these important frames with high motion salience. Experiments on five benchmarks verify the effectiveness of our adaptive sampling over these fixed sampling strategies, and also the generalization power of motion-guided sampling across different backbones, video models, and datasets. 
\vspace{-3mm}
\paragraph{\bf Acknowledgements.} This work is supported by National Natural Science Foundation of China (No. 62076119, No. 61921006), Program for Innovative Talents and Entrepreneur in Jiangsu Province, and Collaborative Innovation Center of Novel Software Technology and Industrialization. The first author would like to thank Ziteng Gao and Liwei Jin for their valuable suggestions.

\appendix

\section{Evaluation on multiple views}
The goal of MGSampler is to provide a holistic sparse sampler and only sample one clip from each video for efficient inference. That is a widely used testing scheme by recent methods in Sth-Sth dataset~\cite{sthsth}. Indeed, multi-view testing could further improve the performance but also increase computaitonal cost. We perform multi-view testing (2 clips and 3 crops)  on our MGSampler in the same manner with TSM\cite{lin_ICCV2019_TSM}, and the result is shown in Table~\ref{tab:multi-view}.

\begin{table}[htbp]
\begin{center}

\footnotesize
\begin{tabular}{ccccc}
\toprule
\textbf{Model} & \textbf{Frames}  & \textbf{Test-Views}  & \textbf{Sampler} & \textbf{Top-1 Acc}  \\ 
\toprule
TSM-R50 & 8 & 1$\times$1 & TSN & 57.9  \\
TSM-R50 & 8 & 1$\times$1 &  \textbf{MG} & 59.8(\textbf{+1.9})   \\
TSM-R50 & 8 & 2$\times$3 & TSN & 61.2    \\
TSM-R50 & 8 & 2$\times$3 &  \textbf{MG} & 62.9(\textbf{+1.7})   \\
\bottomrule
\end{tabular}
\vspace{2mm}
\caption{Multi-view testing on \bf{Something-Something V2}.}
\label{tab:multi-view}
\end{center}
\end{table}
\vspace{-8mm}


\section{Use MGSampler as a clip sampler}
\label{sectionB}
Our MGSampler could be easily adapted to dense clip sampling. The original dense methods samples 8 frames from continuous 32 frames with stride 4. Our MGSampler can adaptively sample a 8-frame clip guided by accumulation curve from the same continuous 32 frames. The results on Sth-Sth V2 are shown in Table~\ref{tab:clip_sampler}, which demonstrates the effectiveness of MGSampler on dense sampling.

\begin{table}[htbp]
\begin{center}
\resizebox{0.45\textwidth}{!}{
\footnotesize
\begin{tabular}{ccccc}
\toprule
\textbf{Model} & \textbf{Frames}  & \textbf{Test-Views}  & \textbf{Clip Sampler} & \textbf{Top-1 Acc}  \\ 
\toprule
SlowOnly-R50 & 8 & 1$\times$1 & fixed stride  & 57.7   \\
SlowOnly-R50 & 8 & 1$\times$1 &  \textbf{MG} & 58.5(\textbf{+0.8})   \\
SlowOnly-R50 & 8 & 10$\times$3 & fixed stride  & 62.1    \\
SlowOnly-R50 & 8 & 10$\times$3 &  \textbf{MG} & 62.5(\textbf{+0.4})   \\
\bottomrule
\end{tabular}
}
\vspace{2mm}
\caption{MGSampler extension as a dense clip sampler. Testing with SlowOnly-R50~\cite{feichtenhofer2019slowfast} on \bf{Something-Something V2}. }

\label{tab:clip_sampler}
\end{center}
\end{table}
\vspace{-8mm}

\section{Results on untrimmed videos}
we extend MGSampler to untrimmed video testing. The results in ActivityNet~\cite{caba2015activitynet} is reported in Table~\ref{tab:anet}. We first perform sparse frame sampling in a TSN-like framework, and our MGSampler is better than TSN by 1.4\%. Then we use MGSampler to perform dense clip sampling as in {\bf Section~\ref{sectionB}} and it is better than standard dense clip sampling by 0.7\%.

\begin{table}[htp]
\begin{center}
\footnotesize
\resizebox{0.44\textwidth}{!}{
\begin{tabular}{ccccc}
\toprule
\textbf{Model} & \textbf{Frames}  & \textbf{Test-Views}  & \textbf{Sampler} & \textbf{Top-1 Acc}  \\ 
\toprule
SlowOnly-R50 & 8 & 1$\times$1 & TSN & 77.4    \\
SlowOnly-R50 & 8 & 1$\times$1 &  \textbf{MG} & 78.8(\textbf{+1.4})   \\
\midrule
SlowOnly-R50 & 8 & 10$\times$3 & 8$\times$8 clip & 80.3    \\
SlowOnly-R50 & 8 & 10$\times$3 &  \textbf{MG}(clip) & 81.0(\textbf{+0.7})\\
\bottomrule
\end{tabular}
}
\vspace{2mm}
\caption{Performance comparison on \bf{ActivityNet 1.3}.}
\vspace{-5mm}
\label{tab:anet}
\end{center}
\end{table}
\vspace{-5mm}

\begin{figure*}[htbp]
    \centering
    \includegraphics[width=1.0\linewidth]{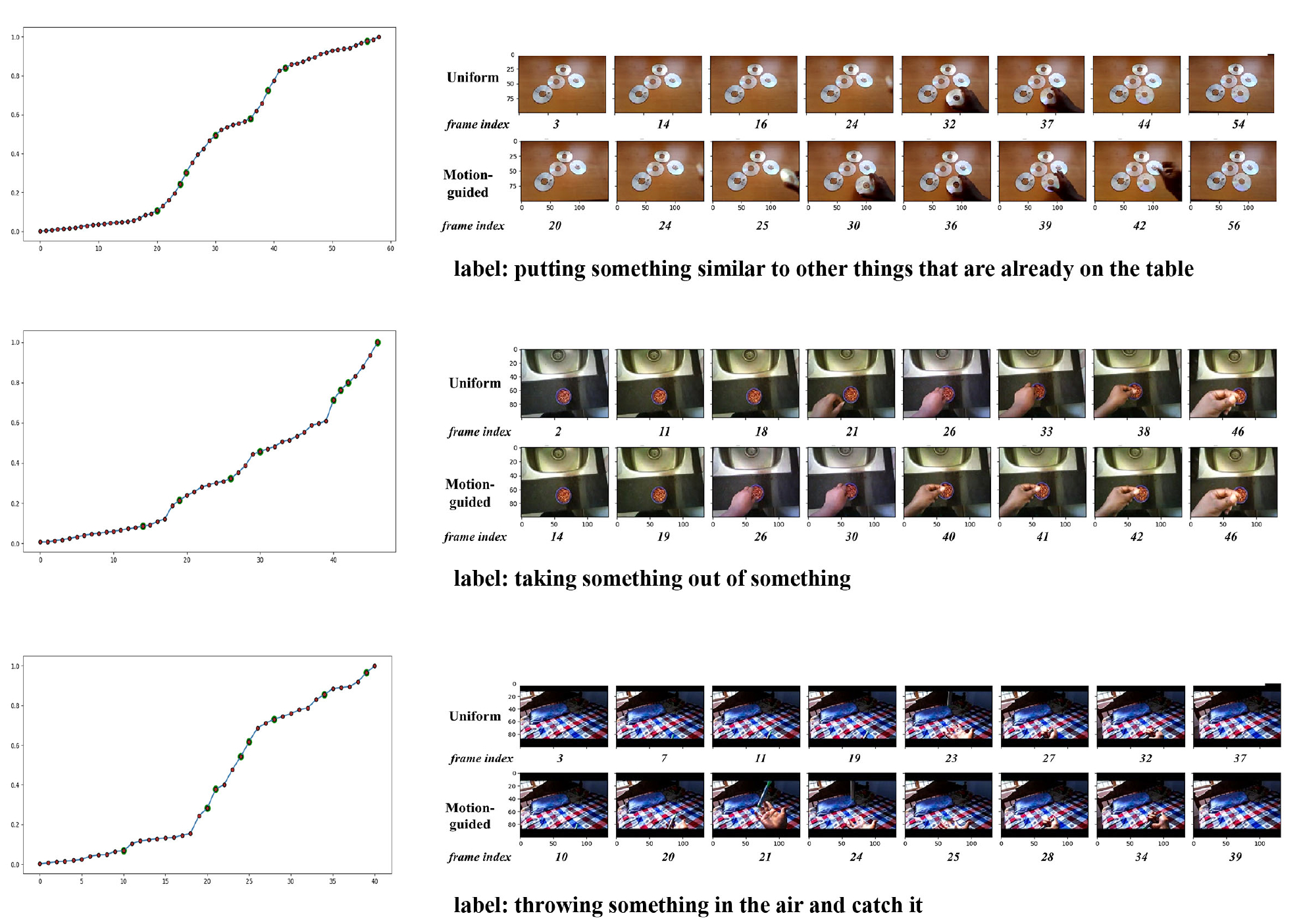}
    \vspace{-3mm}
        \caption{Examples of comparison between uniform sample and motion-guided sample on the Something-Something dataset.}
    \label{s1}
    \vspace{-1mm}
\end{figure*}
\begin{figure*}[htbp]
    \centering
    \includegraphics[width=1.0\linewidth]{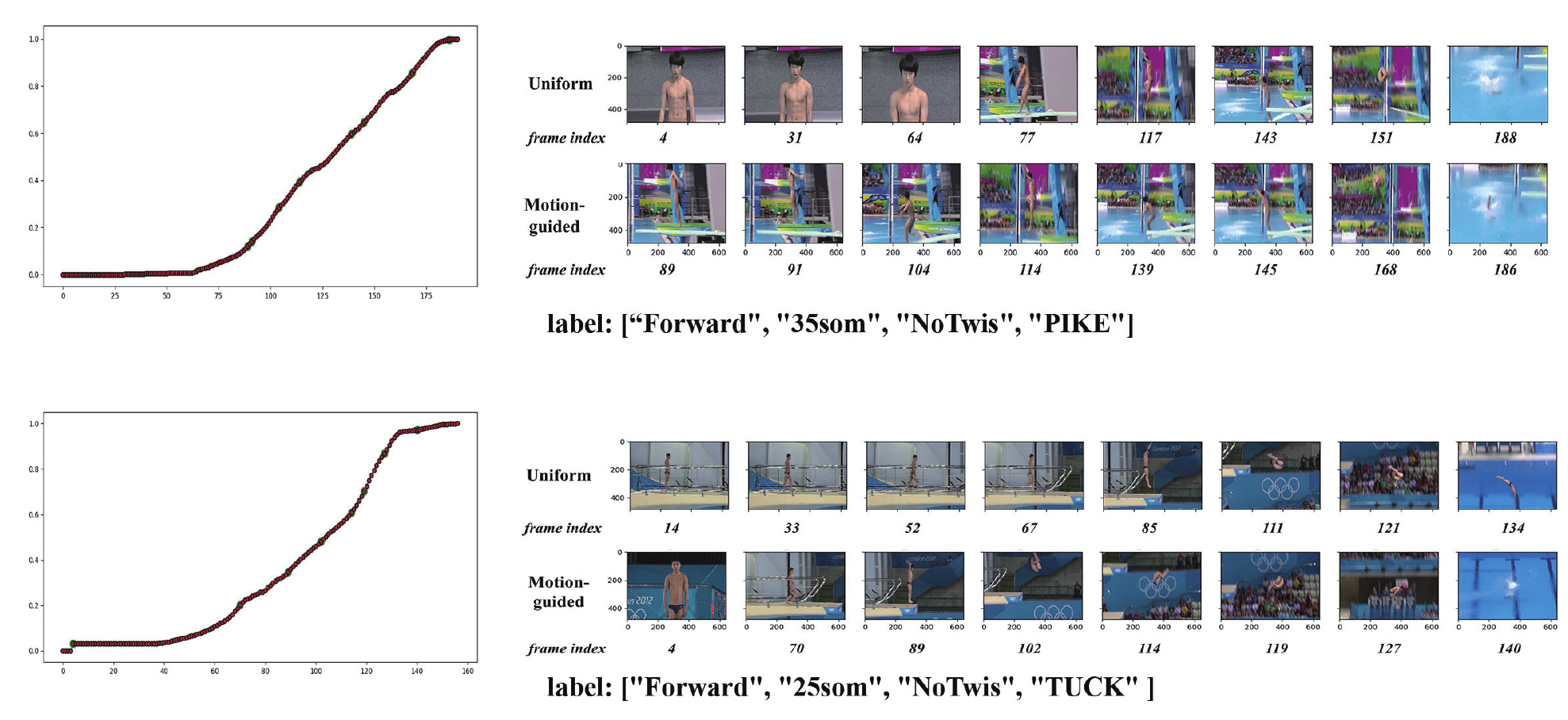}
    \vspace{-3mm}
    \caption{Examples of comparison between uniform sample and motion-guided sample on the Diving48 dataset.}
    \label{s2}
    \vspace{-5mm}
\end{figure*}
\begin{figure*}[htbp]
\centering
\includegraphics[width=1.0\linewidth]{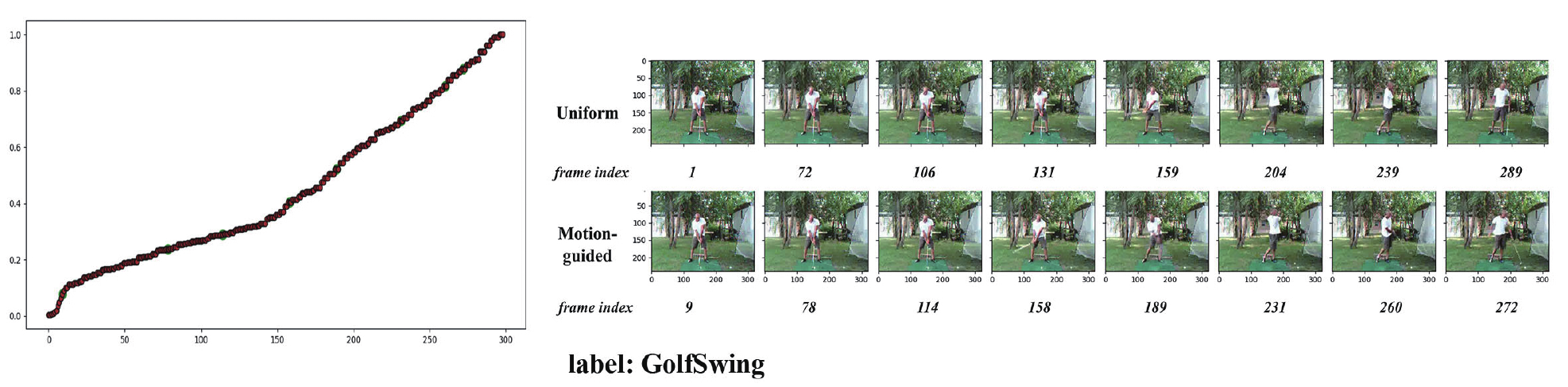}
\vspace{-1mm}
\centering
\includegraphics[width=1.0\linewidth]{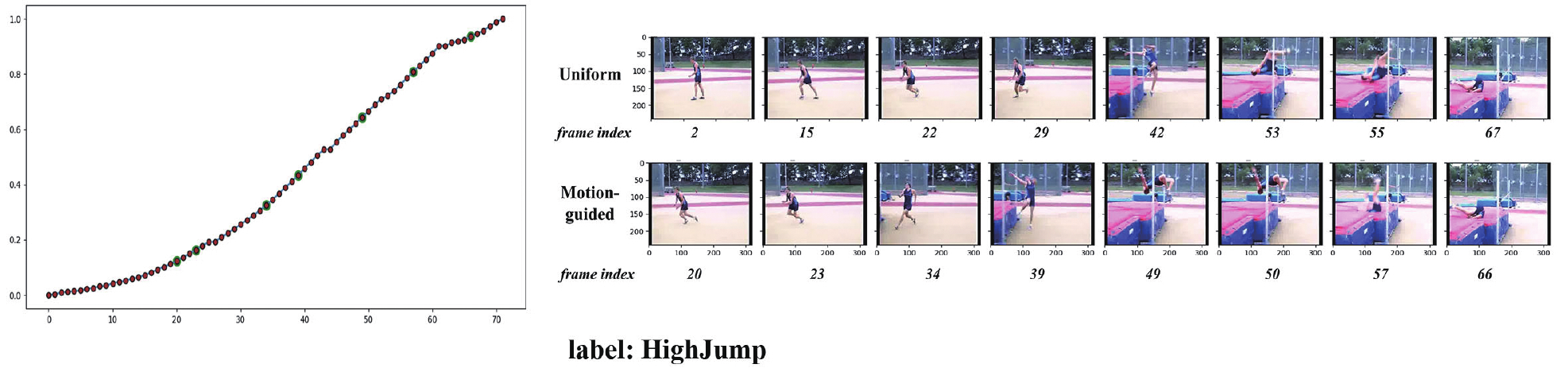}
\vspace{-2mm}
\caption{Examples of comparison between uniform sample and motion-guided sample on the UCF101 dataset.}
\vspace{-2mm}
\label{s3}
\end{figure*}

\begin{figure*}[htbp]
    \centering
    \includegraphics[width=1.0\linewidth]{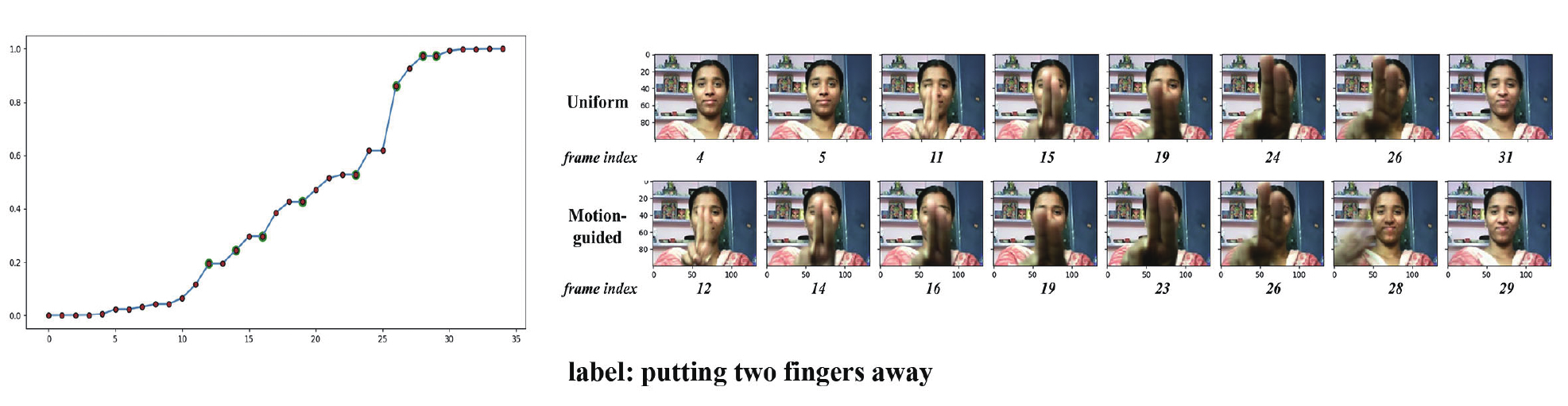}
    \vspace{-3mm}
    \caption{Examples of comparison between uniform sample and motion-guided sample on the Jester dataset.}
    \label{s4}
    \vspace{-2mm}
\end{figure*}
\begin{figure*}[!t]
    \centering
    \includegraphics[width=1.0\linewidth]{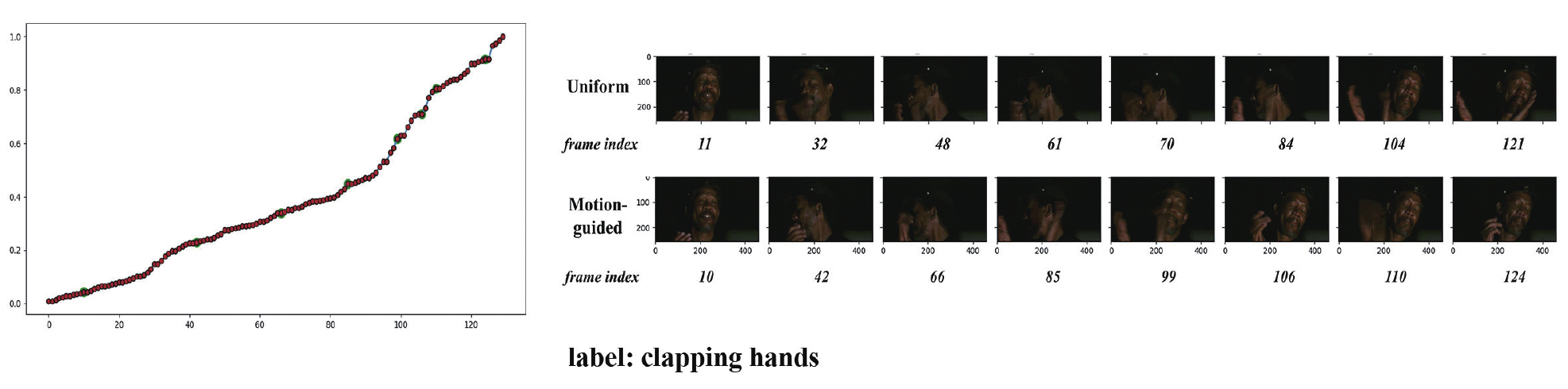}
    \vspace{-3mm}
    \caption{Examples of comparison between uniform sample and motion-guided sample on the HMDB dataset.}
    \label{s5}
    \vspace{-5mm}
\end{figure*}

\section{Visualization analysis}
More examples of comparison between uniform sample and motion-guided sample on Sth-Sth~\cite{sthsth}, Diving48~\cite{diving}, UCF101~\cite{ucf101}, HMDB~\cite{hmdb51}, Jester~\cite{jester} datasets. The left column of Figure~\ref{s1},~\ref{s2},~\ref{s3},~\ref{s4} and ~\ref{s5} is the cumulative distribution motion and the right column is the sampled frames.

{\small
\bibliographystyle{ieee_fullname}
\bibliography{MGSampler}
}

\end{document}